\documentclass[acmtog, authorversion]{acmart}
\makeatletter
\def\@ACM@checkaffil{
    \if@ACM@instpresent\else
    \ClassWarningNoLine{\@classname}{No institution present for an affiliation}%
    \fi
    \if@ACM@citypresent\else
    \ClassWarningNoLine{\@classname}{No city present for an affiliation}%
    \fi
    \if@ACM@countrypresent\else
        \ClassWarningNoLine{\@classname}{No country present for an affiliation}%
    \fi
}
\makeatother
\acmSubmissionID{862}

\usepackage{cleveref}
\usepackage{bm}
\usepackage{amsmath}
\usepackage{comment}
\usepackage{textcomp}
\usepackage{subcaption}

\AtBeginDocument{%
  \providecommand\BibTeX{{%
    \normalfont B\kern-0.5em{\scshape i\kern-0.25em b}\kern-0.8em\TeX}}}

\setcopyright{licensedothergov}
\acmYear{2023} \acmVolume{42} \acmNumber{4} \acmArticle{} \acmMonth{8} \acmPrice{15.00}\acmDOI{10.1145/3592456}

\acmJournal{TOG}
\acmVolume{42}
\acmNumber{4}
\acmArticle{111}
\acmMonth{8}



\begin{document}

\newcommand{\todo}[1]{{#1}}
\newcommand{\new}[1]{{#1}}

\title{BodyFormer: Semantics-guided 3D Body Gesture Synthesis with Transformer}


%
\author{Kunkun Pang}
\authornote{Kunkun Pang and Dafei Qin are joint first authors}
\email{kk.pang@giim.ac.cn}
\affiliation{%
  \institution{Guangdong Key Laboratory of Modern Control Technology, Institute of Intelligent Manufacturing, Guangdong Academy of Sciences}
   \country{China}
}

\author{Dafei Qin}
\authornotemark[1]
\email{qindafei@connect.hku.hk}
\affiliation{%
  \institution{The University of Hong Kong}
   \country{Hong Kong}
}
\author{Yingruo Fan}
\email{yingruo@connect.hku.hk}
\affiliation{%
  \institution{The University of Hong Kong}
   \country{Hong Kong}
}

\author{Julian Habekost}
  \email{julian.habekost@ed.ac.uk}
  \affiliation{%
  \institution{The University of Edinburgh}
      \country{UK}
  }

\author{Takaaki Shiratori}
\email{tshiratori@meta.com}
\affiliation{%
  \institution{Meta Reality Labs Research}
      \country{US}
  }
  
  \author{Junichi Yamagishi}
  \email{jyamagis@nii.ac.jp}
  \affiliation{%
  \institution{National Institute of Informatics}
      \country{Japan}
  }

  \author{Taku Komura} 
\authornote{Corresponding author}
\email{taku@cs.hku.hk}
\affiliation{%
\institution{The University of Hong Kong}
   \country{Hong Kong}
  }
 \affiliation{%
\institution{Research Institute of Electrical Communication, Tohoku University}
   \country{Japan}
  }

\renewcommand{\shortauthors}{Pang and Qin et al.}


\begin{teaserfigure}
\centering
  \includegraphics[width=1.0\textwidth]{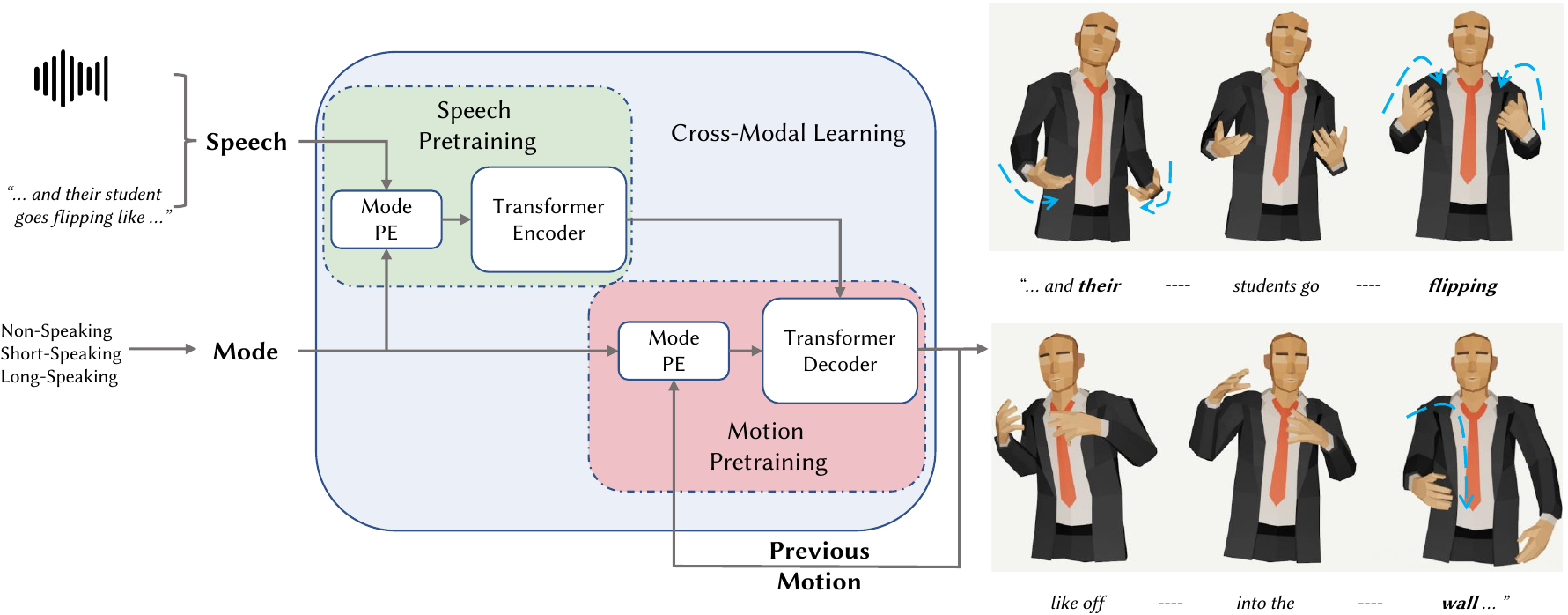}
  \caption{Given an arbitrary input speech, our proposed transformer-based model, BodyFormer, can generate a sequence of vivid 3D body gestures.}
   \label{fig:my_label}
\end{teaserfigure}

\begin{abstract}
Automatic gesture synthesis from speech is a topic that has attracted researchers for applications in remote communication, video games and Metaverse.
Learning the mapping between speech and 3D full-body gestures is difficult due to 
the stochastic nature of the problem 
and the lack of a rich cross-modal dataset that is needed for training. 
In this paper, we propose a novel transformer-based  framework for automatic 3D body gesture synthesis from speech.
To learn the stochastic nature of the body gesture during speech, we propose a variational transformer to effectively model a probabilistic distribution over gestures, which can produce diverse gestures during inference. 
Furthermore, we introduce a mode positional embedding layer to capture the different motion speeds in different speaking modes. 
To cope with the scarcity of data, we design an intra-modal pre-training scheme that can learn the complex mapping between the 
speech and the 3D gesture from a limited amount of data. 
Our system is trained with either the Trinity speech-gesture dataset or the Talking With Hands 16.2M dataset. The results show that our system can produce more realistic, appropriate, and diverse body gestures compared to existing state-of-the-art approaches.

\end{abstract}
\maketitle

\section{Introduction}

Automatic synthesis of 3D body gestures from speech is a problem that researchers in psychology, computer graphics and computer vision have been tackling. Most classic approaches are either rule-based  where the corresponding gesture for each context is carefully designed based on observation, or make use of low-level features of speech such as prosody to produce movements that are well synchronized with the speech. We wish to go beyond such carefully designed architectures and learn a mapping from speech to body gesture automatically from a mid-size dataset.

However, there are various difficulties in learning the task of body gesture synthesis from speech. First of all, this is a cross-modality learning problem that requires a significant amount of training data for producing a proper mapping. Unlike training deep language models (e.g. BERT ~\citep{devlin-etal-2019-bert}) and vision models 
(e.g., ViT~\citep{dosovitskiy2021an}), collecting a large amount of training data, which is motion capture data synchronized with speech data, may not be practical. Secondly, the correlation between speech and gesture is rather weak. Regressing body gestures from low-level speech features may easily fail due to the ambiguity of the mapping. 
Thirdly, gestures are usually related to either the high-level context of the speech contents or the mode of the current conversation, such as moving the hands and fingers to express the adjective of a subject, nodding while agreeing and side-stepping while listening, which are difficult to be learned unless high-level semantic information and mode label is involved.

In this paper, we present \textit{BodyFormer}, a novel transformer-based framework for speech-driven 3D body gesture synthesis, as depicted in \cref{fig:my_label}. 
Since human body movement is a stochastic process, and there is significant ambiguity between speech and motion, we propose a generative transformer architecture with variational inference to synthesize 3D body gestures. To cope with the difficulty of cross-modality learning, we use a transformer encoder-decoder framework with early and intermediate modality fusion. More specifically, the transformer encoder encodes low-level and high-level information, where the low-level features preserve the audio frequency and the high-level semantics strengthens the weak correlation between the full-body motion and the speech signal. 
Such information is fused before being fed into the transformer encoder. Then, the transformer decoder with a multi-head attention scheme can merge both speech and the previous motion modality from the intermediate layers to produce the speech-related gesture.
To cope with the data-hungry nature of cross-modal transformer training \cite{khan2021transformers},
we devise an intra-modal pre-training strategy, which alleviates the difficulty of cross-modal learning for speech and body gestures.
Additionally, to learn the motions of different speeds in various speaking modes, e.g. not-speaking, short-speaking, and long-speaking, we propose a mode positional embedding layer to learn mode-dependent period parameters to capture such differences.

We conduct comprehensive experiments to evaluate our system using the Trinity speech-gesture dataset \cite{IVA:2018}
and TalkingWithHands 16.2M dataset~\cite{Lee_2019_ICCV}.  
Vivid gestures that represent the semantic expression of the speaker can be generated from the speech. We conduct quantitative and qualitative comparisons with the baselines and show that our model outperforms the state-of-the-art. Our system can be applied to controlling virtual presenters, characters in films, and AI agents in virtual environments.

The contribution of the paper is summarized as follows:
\begin{itemize}
    \item A transformer framework for synthesizing 3D body gestures from speech, which can produce expressive movements where the gestures are in sync with the speech and also contextually related,   
    \item introduction of mode positional embedding that helps the transformer to capture the different motion speeds in different speaking modes and
    \item an intra-modal pre-training scheme that enables our transformer framework to learn from a relatively small motion and speech dataset.
\end{itemize}

\section{Related works}

\paragraph{Weak Correlation between Signals}
The correlation between speech and gesture has been a long-term interest in  psychology~\cite{mcneill1992hand,cassell1999speech,wagner2014gesture}. Kendon~\shortcite{kendon1972some} analyzes the synchronization of the speech and gesture and finds that the gesture appears even earlier than the speech.  McNeill~\shortcite{mcneill1992hand} insists that gesture and speech are occurring from a common source. De Reiter et al.~\shortcite{de2012interplay} claim that gesture and speech are complementary to each other to convey the speaker's intention.

 \paragraph{Trade-off Between Quality and Scale} While psychologists are keen on discovering the relationship between signals, speech-driven gesture synthesis is also appealing in the computer graphics and vision community. Not only does signals' correlation need to be well understood, but the quality of generated poses matters. Due to the high cost of the motion capture process, collecting large-scale, high-quality data is impractical. On the contrary, noisy key points could be extracted from videos via pose estimation techniques. This introduces a trade-off between data quality and the dataset scale, which leads to a different research focus. A number of studies~\cite{ginosar2019learning,ahuja2020style,EMNLP,Templates} are aimed at generating 2D body gestures from speech. These 2D gesture generation approaches are not directly applicable to 3D scenarios, for instance, controlling a virtual character in films or video games. Recently researchers~\cite{10.1145/3414685.3417838,HA2G,habibie2021learning, zhu2023taming} shifted their focus to learn the estimated 3D poses from in-the-wild videos, such as the 144-hours TED \cite{AttentionSeq2Seq}, and 251-hours PATS \cite{ginosar2019learning,ahuja2020style} datasets. However, the estimated 3D poses are still noisy and lack joint rotation information. This results in generating ambiguous and less accurate poses, which are hard to feed into the graphics pipeline. Other researches~\cite{kucherenko2020gesticulator, DBLP:journals/cgf/AlexandersonHKB20, rhythm_sigasia2022_ao} focus on learning more accurate body gestures from relatively smaller motion capture datasets synchronized with speech, such as 4-hours Trinity \cite{IVA:2018}, and 20-hours TalkingWithHands 16.2M \cite{Lee_2019_ICCV} dataset. 
 It is noted that those motion capture datasets are 10-30 times smaller than the in-the-wild video datasets. Hence, it is challenging to train a data-hungry model on these relatively smaller datasets.

\paragraph{Deep Motion Generation Methods} The recent deep learning methods typically utilize convolution~\cite{ginosar2019learning,ahuja2020style}, feed-forward~\cite{kucherenko2019analyzing,kucherenko2020gesticulator} and Recurrent Neural Networks (RNN)
~\cite{Audio2body,AttentionSeq2Seq}
to learn the mapping from speech to gestures. For instance, Ginosar et al.~\shortcite{ginosar2019learning} propose a CNN-based framework to predict gestures from audio. Ahuja et al.~\shortcite{ahuja2020style} construct a Temporal Convolution Network (TCN) to formulate a stylized gesture space. Kucherenko et al.~\shortcite{kucherenko2019analyzing} explore different types of audio features. Their follow-up work \cite{kucherenko2020gesticulator} takes both acoustic and textual features as the input of a feed-forward neural network for gesture generation. The probabilistic method~\cite{li2021audio2gestures} employs a conditional VAE to generate diverse body gestures. On the other hand, RNN-based models are also wildly used in speech-driven gesture generation~\cite{Audio2body,ferstl2019multi,AttentionSeq2Seq,ahuja2020style,HA2G, Bhattacharya_speech2affective}. Shlizerman et al.~\shortcite{Audio2body} train an LSTM model to learn the correlation between audio features and body key points. Yoon et al.~\shortcite{AttentionSeq2Seq} build a GRU-based model for gesture generation, where the model is trained on the TED dataset. \new{Wang et al.~\shortcite{Wang_integrateT2SG} propose to improve the motion quality by jointly synthesizing speech and gestures from the text in an integrated LSTM architecture. Liu et al.~\shortcite{liu2022beat} propose a cascaded LSTM and MLP by integrating emotion, speaker identity, and style features for motion synthesis. Liu et al.~\shortcite{Liu_DisCo} further improve the LSTM generator with a content-balanced distribution to tackle the imbalance learning problem caused by fewer frequent motion sequences. Ao et al.~\shortcite{rhythm_sigasia2022_ao} improve the audio-gesture synchronization by cutting inputs into short clips according to beats. They incorporate gesture lexeme encoder modules and disentangle audio features to build a better mapping between speech and gesture. Besides, Habibie et al.\shortcite{Habibie_gesturematching22} search for the most plausible audio-motion sequence from the database with the K-nearest neighbors and refine the motion with the convolution network.} However, these previous deep models can only capture limited contextual information, e.g., RNN-based models inevitably `forget' the past information~\cite{Schmid-lstm-1999, dai-etal-2019-transformer}. Although several methods~\cite{10.1145/3414685.3417838,kucherenko2020gesticulator} extract BERT~\cite{devlin-etal-2019-bert} features from the text, they do not consider the long-term audio context.

\paragraph{Transformer based models} Compared to CNN and RNN based models, the transformer model~\cite{NIPS2017_3f5ee243} is relatively less explored in the audio-driven motion synthesis. \new{Saeed et al.~\shortcite{Saeed_zeroEGGS} present a variational transformer for encoding style information, whereas they adopt recurrent networks to model motion generation from both speech and style.}
Valle-P\'{e}rez et al.~\shortcite{Transflower}
and Li et al.~\shortcite{li2021learn}
propose generative transformer approaches with normalizing flow for dancing motion synthesis from music. However, using the low-level audio feature is less likely to produce contextual movement in our task. 
Unlike Transflower~\cite{Transflower,li2021learn}, Text2gestures~\cite{bhattacharya2021text2gestures} use text and a deterministic transformer to generate body gestures without the low-level audio features. Audio features complement text information, such as pitch, tempo, and loudness, which may influence body gestures. Therefore, we exploit both low-level audio features and the text’s contextual semantic features.

In this paper, we present a novel system for synthesizing motion from speech, which incorporates state-of-the-art features from natural language processing, speech processing, and human motion synthesis. In contrast to existing methods, which overlook the distinct body language patterns during different speaking modes, our proposed system for synthesizing motion from speech incorporates a mode positional embedding method to capture such variations, which enhances the vividness of motion. Besides, the proposed intra-modal pre-training helps prevent the model from overfitting to the training data.
\section{BodyFormer}

Generating the body gesture from the speech is an inherently ambiguous problem where different realistic gestures could match to the same speech. To cope with this issue, we propose a generative model where the system samples noise per sequence to produce variations of output movements. In this section, we first introduce the data pre-processing step, and then, the proposed architecture and its training scheme.
\subsection{Pre-processing}
Raw speech data contain both low-level and high-level features. The low-level features have syllables and phonetic information, and the high-level features provide semantic contextualized information. We use the librosa \cite{librosa} library to transform the audio signal to mel-spectrogram with 27 channels to extract such low-level features \cite{DBLP:journals/cgf/AlexandersonHKB20}. 
For computing high-level features based on the context, we first  
apply Automatic Speech Recognition (ASR) to extract sentences from speech, and then compute the BERT feature \cite{devlin-etal-2019-bert} of each word. Since the size of the BERT feature is much larger than that of the mel-spectrogram feature,
we reduce the dimension of the BERT features to 32 by applying PCA. The mel-spectrogram and BERT features are concatenated to form the  
speech feature at frame $t$, represented by $\bm x_t\in \mathbb{R}^{59}$. 


We construct an equal time-length representation of the audio and words whose temporal resolution is the same as that of the motion (20Hz). For the audio feature, if the sound sampling rate is 48KHz in the original data, we extract 2400 features between two sample points (48K/20 = 2400) and concatenate them to be
the audio feature of that frame. The BERT feature of the corresponding word is set as the high-level feature of that frame.  For the frames where the sound level is low, the word feature is set to zero vectors.   
The speaking mode label, i.e., not-speaking (NS), short-speaking (SS), and long-speaking (LS), of the speaker is  computed automatically by measuring the time length of the sentence and added as a feature (see appendix \ref{apdx_mode} for more details). 



\begin{figure}[t]
	\centering
	\includegraphics[width=0.48\textwidth]{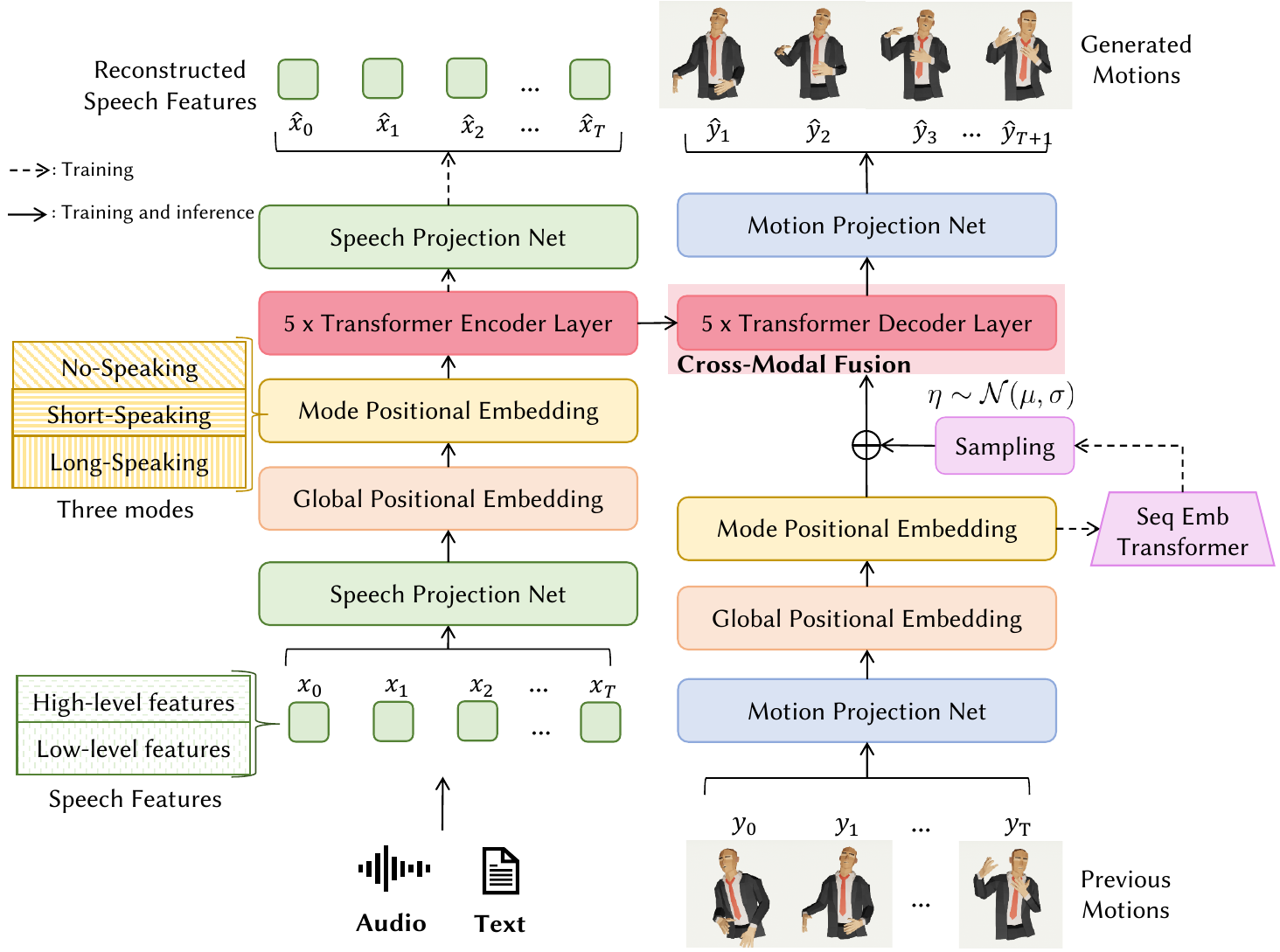}
	\caption{The architecture of the proposed model, BodyFormer. Our model takes both low-level and high-level speech features as input and generates a sequence of realistic 3D body gestures in an auto-regressive manner. The embedding layer consists of global positional embedding (GPE) and mode positional embedding (MPE). GPE captures the global order information for each sequence, whereas MPE depicts the local time information for every mode (NS, SS and LS) by learning the mode-dependent period parameters. Besides, a sequence embedding transformer is applied to learn the motion distribution conditioned on the previous motions. During testing, the encoded previous motions are summed with the noise sampled from the learned distribution for variational inference.}
	\label{fig:arch}
\end{figure}

To representation motion, we use the 6D representation for joint rotation~\cite{zhang2018mode,Zhou_2019_CVPR}. 
The pose representation is the concatenation of all joint rotations $\bm y_t=[\bm z_{t,1}, \cdots, \bm z_{t,J}]\in\mathbb{R}^{6J}$ where $\bm z$ is the joint rotation and $J$ is the number of joints. In this paper, the system will output the upper body and finger pose for the Trinity speech-gesture dataset \cite{IVA:2018} and the full body pose for the Talking with hands 16.2M dataset \cite{Lee_2019_ICCV}.



\subsection{Architecture}

As shown in the \cref{fig:arch}, our proposed system is composed of five sub-networks, including element-wise projection nets, embedding layer, transformer encoder-decoder, sequence embedding transformer, and sampling layer. 

\paragraph*{Projection net} The element-wise projection net aims to map the data to the desired dimensions. More specifically, each projection net consists of 2 fully connected layers with GELU activation functions. As shown in the \cref{fig:arch}, both speech and motion features will be sent to their corresponding projection net and transformed to 512 dimensions. Similarly, the transformers' outputs are projected to their desired output size with a linear layer: We call the linear layer the projection layer for consistency.



\paragraph*{Transformer} Similar to the vanilla transformer~\cite{NIPS2017_3f5ee243}, we use the same encoder-decoder framework to model the speech-driven motion generation task. The encoder takes speech $\bm x =(\bm x_1, \bm x_2, \dots, \bm x_T)$ as input and encodes it into contextualized representation $\bm h_{1\to T}=f(\bm x_{1\to T}')$ without masks, where $\bm x'$ is speech feature after the speech projection net and embedding layer. The decoder will receive both previous motions $\bm y_{<t}=(\bm y_1,\bm y_2,\dots,\bm y_{t-1})$, the encoded speech information, and a lower-triangulated mask to produce the current pose:
\begin{align}
	\bm y_t = g(\bm y_{<t}, f(\bm x')),
\end{align}
where $f$ and $g$ denote the encoder and decoder, respectively. The triangulated mask ensures the decoder only uses the past motion without cheating on using the future pose information. Unlike the bidirectional model \cite{li2021learn}, our decoder is trained with shifted-by-1 supervision such that the decoder can learn the pose prediction in an auto-regressive manner.

\paragraph*{Embedding layer} 
The embedding layer is composed of Global Positional Embedding and Mode Positional Embedding. 
As stated in \cite{NIPS2017_3f5ee243}, the transformer lacks recurrence and convolution to make use of the order of sequence. They propose \emph{Global Positional Encoding (GPE)}, which provides time and order information to enable the model to process the entire sequential data and preserve continuity. Here, we follow \cite{wang2021on} and use a learnable sinusoidal function:
\begin{align}
	GPE(t)=[\sin(\omega t), \cos(\omega t)], t=\{0,1,\dots,T\}.
\end{align}
where $\omega$ is the learnable period parameter.


Apart from this, we hypothesize that the motion speed tends to be different in different modes. The global positional embedding layer  may potentially average the movement speed for all modes since modes could randomly occur in a long sequence. Besides, the period parameter of GPE is identical in either speaking or not speaking. This may lead the generated motion to be less active during speaking, and not salient during not-speaking. Thus, we present an additional \emph{Mode Position Embedding (MPE)} to learn mode-dependent period parameters $\omega_m$, 
\begin{align}
	MPE(m,t')&=[\sin(\omega_m t'), \cos(\omega_m t')], \nonumber \\
	&t'=\{0,1,\dots,T'\}, 
\end{align}
where $m$ is a speaking mode label (NS, SS or LS) and $T'$ is the length of a detected speaking mode.
One noticeable difference between GPE and MPE is the time horizon. GPE describes global order information for every sequence with a fixed length $T$, whereas MPE has dynamic time horizons $T'$ based on the length of the corresponding mode. 
Each time step $t'$ of MPE depicts the local time information for every speaking mode individually. 

Finally, we combine both local and global embedding to compose the embedding layer 
$Embed(x,t,t',m) = \text{LayerNorm}(c*(x+GPE(t))+MPE(m,t'))$
, where $c=\sqrt{512/3}$ is a constant to strengthen the information from the raw inputs and the global time.

\paragraph*{Variational Inference}
The system is also expected to generate various motion sequences based on the states of previous motion and corresponding speech. Here, we propose to model the posterior distribution with a sequence embedding transformer that encodes the sequence into a single vector. The prior distribution will be modeled as a learnable multivariate distribution $\bm \eta\sim\mathcal{N}(\bm \mu, \bm \sigma)$. In training time, we train both sequence embedding transformer and the multivariate normal distribution simultaneously. In test time, the sequence embedding transformer will be disabled, and we will randomly sample from the learned multivariate normal distribution.

Specifically, our sequence embedding transformer uses the same architecture as the set transformer~\cite{Settrans}. We use a shallower set transformer \new{$\phi(\cdot)$} with only two multi-head attention blocks (MAB) and a pooling layer with multi-head attention (PMA). The multi-head attention block is identical to the vanilla transformer encoder and decoder:
\begin{align}
	MAB(Q,K,V)=\text{LayerNorm}(H+\text{FF}(H)),\\
	 H=\text{LayerNorm}(Q+\text{Multihead}(Q,K,V)),
\end{align}
where FF, LayerNorm, and Multihead represent the fully connected layer, layer normalization, and multi-head attention, respectively. Q, K and V indicate queries, keys and values, which are the inputs of multi-head attention. In our case, they all come from the previous mode position embedding layer. Instead of applying an element-wise average or maximum of the instances, we pool the data by multi-head attention. To apply multi-head attention as a pooling function, PMA has a learnable seed vector $S\in \mathbb{R}^{k\times d}$. Let us denote $H_{set}\in\mathbb{R}^{n\times d}$ as the set of features:
\begin{align}
	\text{PMA}_k(H_{set})=MAB(S, FF(H_{set})).
\end{align}
Then, the output of the PMA will be a set of $k$ items. Note that we use one seed vector ($k=1$) for this task.

\subsection{Objective Functions}
To train this system, we introduce three loss functions for learning the relation between speech and motion: \emph{Joint Prediction Loss}, \emph{Magnitude Loss}, and \emph{KL Divergence}.

\paragraph{Joint Prediction Loss:} Firstly, we introduce an MSE loss to learn to predict the corresponding pose for every time step. This loss is mainly to learn the correlation between speech and the joints' spatial information:

\begin{align}
	L_g=\frac{1}{T}\sum_{t}^{T}(\hat{\bm y_t}-\bm y_t^*)^2,
\end{align}
where $\hat{\bm y_t}$, $\bm y_t^*$ are the predicted and ground-truth poses, respectively.

\paragraph{Magnitude Loss:} We further introduce a magnitude loss to learn temporally coherent motion. This loss is used to encourage the generated motion changes to be similar to the ground truth. This loss can smooth the generated motion and make it look more vivid: 
\begin{align}
	L_m=\frac{1}{TJ}\sum_T\sum_J \left(||\hat{\bm{z}}_{t,j}-\hat{\bm{z}}_{t-1,j}||_F-||\bm{z}_{t,j}^*-\bm{z}_{t-1,j}^*||_F \right)^2,
\end{align}
where $\hat{\bm{z}}_{t,j}$, $\bm{z}_{t,j}^*$ are the components of  $\hat{\bm y_t}$, $\bm y_t^*$, respectively.

\paragraph{KL Divergence} Additionally, we minimize the KL divergence between the vectorized sequential distribution and a learnable multivariate normal distribution:
\begin{align}
	L_{KL}= KLD(q(\eta|\phi(\bm y))|p(\eta))
\end{align}
where $q(\eta|\phi(\bm y))$ is the posterior distribution from the output of the sequence embedding transformer and $p(\eta)=\mathcal{N}(\bm \mu,\bm \sigma)$ is the learnable multivariate normal distribution with a diagonal covariance matrix. This loss is introduced to encourage the model to be less deterministic, as well as to serve as regularization that prevents overfitting.

Finally, we combine all three terms together:
\begin{align}
	\ell = \lambda_1 L_g + \lambda_2 L_m + \lambda_3 L_{KL}
\end{align}
where we set $\lambda_1=1, \lambda_2=0.3$ to weight the two different loss functions, and $\lambda_3$ is an additional parameter to prevent the posterior collapse, which is controlled by the Cyclical Annealing Schedule \cite{fu-etal-2019-cyclical}. In our case, our $\lambda_3$ is uniformly increased from 0.2 to 3 in 10 epochs and then reset to 0.2 again.  

\subsection{Learning Strategy}

The nature of the body gesture synthesis from the speech is a cross-modal problem; tackling such a problem is one of the most critical parts of producing high-quality human gestures. Due to the difference between the distribution of speech and motion, we simplify the problem by pre-training the speech transformer encoder and motion transformer decoder individually so that the cross-modal learning will start from a reasonable intra-modal manifold.

\paragraph*{Intra-modal pre-training} We pre-train the encoder and the decoder by optimizing Masked Speech Modelling (MSM) and Masked Motion Modelling (MMM). Both MSM and MMM are similar to Masked Language Modelling (MLM), which estimates the missing speech and pose based on the available frames, respectively. Such unified intra-modal pre-training tasks benefit the encoder and decoder as the pre-trained embedding with MSM and MMM can be helpful for successive tasks in either bidirectional or autoregressive.

Unlike pre-training BERT \cite{devlin-etal-2019-bert} and ViT \cite{dosovitskiy2021an}, the size of the 3D motion capture and speech dataset is much smaller than the datasets for pre-training in the natural language processing \cite{conf/iccv/ZhuKZSUTF15} and computer vision community~\cite{deng2009imagenet}. This limited size of motion and speech training samples makes the pre-training even more challenging if we use the same hyperparameter from BERT. Thus, we choose an easier hyper-parameter without mask token during the pre-training phase. Similar to BERT \cite{devlin-etal-2019-bert}, we modify $20\%$ of the speech data or motion data, where $10\%$ of the modified data are replaced as noise drawn from a normal distribution, and the rest will be set as zero. 

The transformer encoder and decoder are trained separately in this phase. Unlike the speech transformer encoder, the motion transformer decoder has both the multi-head self-attention and the multi-head attention block. During pre-training, we disable the multi-head attention block of the transformer decoder, and the data flow bypasses the next corresponding block. Note that the data flow does not go through the sequence embedding transformer and the learnable prior distribution.

\paragraph*{Cross-modal learning} After pre-training both the speech encoder and motion decoder, we perform cross-modal learning. Specifically, multi-head attention learns to address the cross-modal learning of speech and motion. We follow the same routine as the vanilla transformer decoder framework \cite{NIPS2017_3f5ee243} that uses the multi-head self-attention module to encode the past motion and the multi-head attention module to get the information from speech features. The transformer decoder receives the encoded speech features from the transformer encoder and the embedded pose features from motion embedding to compute the next pose in an auto-regressive fashion where their correlation are learned. All system parameters are updated in this phase by minimizing the objective functions. In addition, the sequence embedding transformer and the multivariate normal distribution are also learned in an end-to-end manner.

\subsection{Implementation Details}

Training a deep transformer is often challenging due to the strong capability of the architecture, which leads the neural network to fall into poor local minima, i.e. overfit to the training data and failure to generalize to real-world data. We use several tricks to train such a complex model successfully.

\paragraph*{Warmup Scheduler} Similar to the tricks for training transformer \cite{NIPS2017_3f5ee243, devlin-etal-2019-bert}, we follow the same cosine warmup learning rate scheduler. This warmup learning rate scheduler has the benefit of stabilizing the training process from a relatively better initialization than random by tuning the attention module at the early learning stage. For all experiments, we set the warmup step to 400 epochs.

\paragraph*{Spec Augmentation} 
To train our model with a relatively small body gesture-speech dataset, we use spec augmentation \cite{Park2019SpecAugmentAS}. This increases the size of the training data and improves the generalization performance. For simplicity, we randomly zero out consecutive frames of audio or words, respectively, with the range of $0\%-20\%$.

\paragraph*{Dropout Annealing} Training the transformer or other time-series models in an auto-regressive manner may easily fail in the test time. Since the pose between the current and the next step is often similar, the link between speech and pose is unapparent. This leads to the deep transformer or other auto-regressive model ignoring the speech information's importance and overfits to the previous poses. In \cite{DBLP:journals/cgf/AlexandersonHKB20}, they apply a frame-level dropout on previous motions to strengthen the connection from speech. However, using the same frame-level dropout to the input motion may occasionally fail to learn the reasonable attention heatmap. The transformer may overfit and ignore the speech around the current frame. In our case, we apply a dropout annealing on the motion input where the dropout probability will be annealed from 100\% to 60\%. This will force the multi-head attention to capture as much information as possible during the warmup phase.

\begin{figure*}[t]
	\centering
	\includegraphics[width=0.95\textwidth]{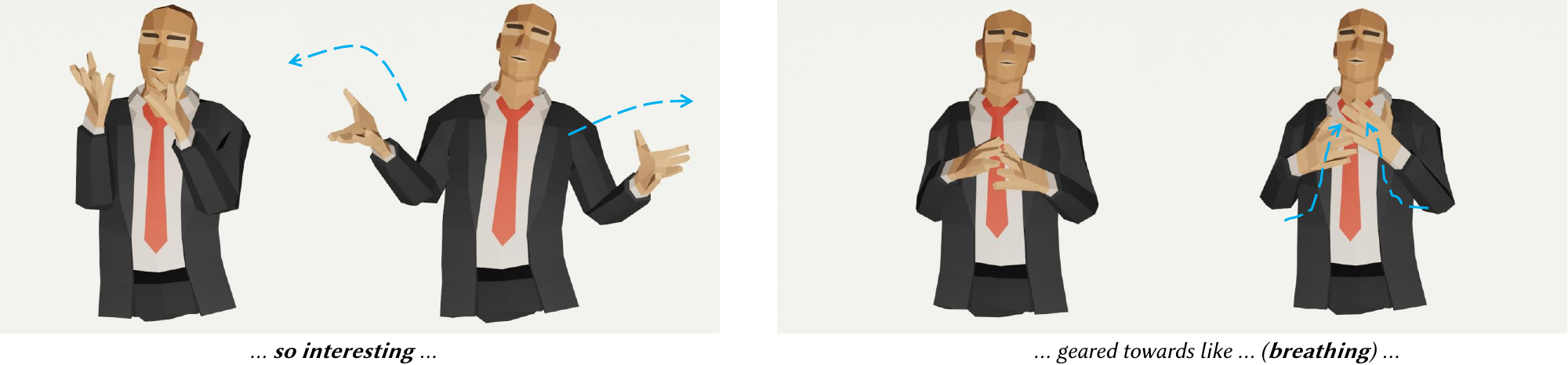}
	\caption{Gesture synchronization. \textit{Left:} The character spreads the arms to express the amount of interest it has felt. \textit{Right:} The character takes a breath and holds the arms to the chest so as to reflect possession of a notion for which an immediate vocabulary is not available.
	\label{fig:sointeresting}}
\end{figure*}

\section{Experiments and Results}

We compare our model with various baseline models by qualitative case studies and quantitative metrics. Since human gestures are subtle to evaluate, we conduct extensive user studies to demonstrate the quality of our generated gestures.

\subsection{Experiment Design}

We train and evaluate our proposed system on two datasets: Trinity Speech-Gesture Dataset~\cite{IVA:2018} and Talking With Hands 16.2M (TWH) dataset~\cite{Lee_2019_ICCV}. The Trinity dataset is often used for the co-speech body gesture synthesis task. 
We use the training and test splits provided by the 2020 GENEA body gesture challenge~\cite{kucherenko2021large}.
Regarding the TWH dataset, we use the deep capture set 
that is composed of full-body motion during the conversations of two subjects.  As only one person's motion is available in most conversations, and the finger motion may occasionally fail to capture, we use the mocap data of the single person with reasonable finger motion for training.


We evaluate the performance with the Trinity and TWH datasets. On the Trinity dataset, we compare our model (BodyFormer) with StyleGestures (StyleGest)~\cite{henter2020moglow, DBLP:journals/cgf/AlexandersonHKB20}, Trimodal~\cite{10.1145/3414685.3417838}, Gesticulator (Gest) \cite{kucherenko2020gesticulator} and Aud2Repr2Pose (A2R2P)~\cite{kucherenko2019analyzing}. We also compare with a baseline where the mocap data is paired with a random audio sequence. We call this baseline as \textit{Unmatched competitor} (Unmatch). On the TWH dataset, we compare BodyFormer with Trimodal, Gesticulator, and Unmatch.

Our model is trained on PyTorch \cite{NEURIPS2019_9015} using AdamW optimizer~\cite{loshchilov2018adamw} with hyper-parameters $(\beta_1 , \beta_2) = (0.9, 0.999)$. The learning rate is set to $10^{-4}$ for pre-training and $10^{-5}$ for cross-modal learning. The weight decay is set to $10^{-2}$ for the pre-training and cross-modal learning phase. Both phases are run on 4 $\times$ Nvidia 2080Ti GPUs, and trained with (50, 1500) epochs with a batch size of 32, respectively. The training takes 7 days in total.

\subsection{Qualitative Evaluation}
Our model can produce smooth and realistic gestures that are in sync with the speech.  
We highlight some of the interesting motion that are produced by our system.  
The model trained with the Trinity dataset produces rich, active gestures that contextually match the speech.  
When the speaker is describing his experience, he mentions "It was SO interesting...", and spreads his arms in sync with the word "so", to express the amount of interest that the speaker felt (see \cref{fig:sointeresting}, left).  
In another example, the speaker mentions ``kind of geared towards LIKE...''.  When he speaks the word ``like'', the hands are retained closer to the chest so as to reflect possession of a notion for which an immediate vocabulary is not available, which often appears in real-life (see \cref{fig:sointeresting}, right).  

\begin{figure*}
\includegraphics[width=\textwidth]{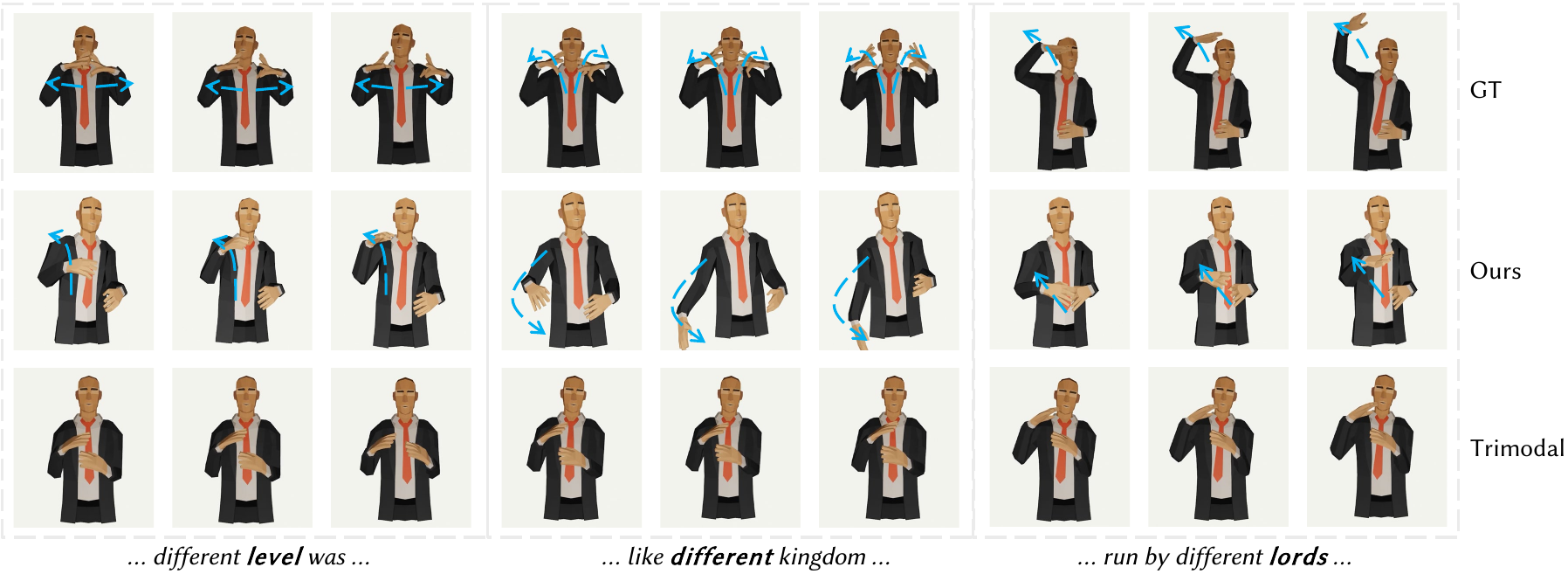}
\caption{\label{11-competitors} \textit{Left:} To emphasize the word \textit{\textbf{level}}, our character raises his right hand  while the ground truth opens both hands. \textit{Middle:} Our character also has an action, though distinct from the ground truth, to express the meaning of \textit{\textbf{different}}. \textit{Right}: The character raises his hands in both our model and the ground truth. The discrepancy is that the ground truth is exaggerated. However, in Trimodal, the character always puts his hands in front of the chest. }
\end{figure*}

\begin{figure*}
\includegraphics[width=\textwidth]{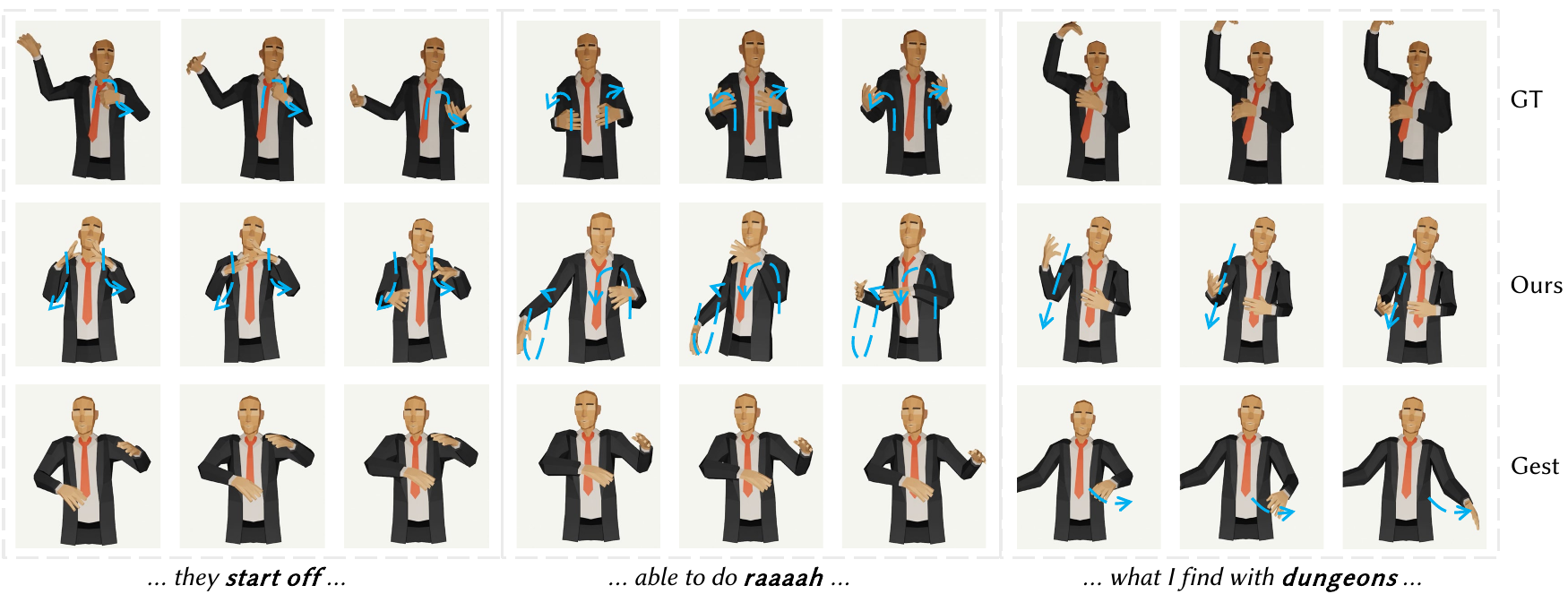}
\caption{ \textbf{Left:} GT and Ours emphasize \textbf{\textit{start}} and \textbf{\textit{off}}, respectively.  \textit{Middle:} To accompany with the onomatopoeia \textbf{\textit{raaaah}}, our model produces a specific gesture. \textit{Right:} GT picks a unique gesture to exemplify the word \textbf{\textit{dungeons}}. Ours and Gest both move one hand to emphasize.}
\label{fig:competitor-style}
\end{figure*}

In \cref{11-competitors}, we visually compare the quality of gestures generated from BodyFormer, Trimodal and the ground truth. BodyFormer outputs either similar gestures to the ground truth (\cref{11-competitors}, left, right), or different but semantically correlated gestures (\cref{11-competitors}, middle). However, the gestures from Trimodal are static, fail to emphasize the speech. In \cref{fig:competitor-style} we evaluate the semantics of the gestures from BodyFormer, StyleGestures and the ground truth. While the gestures of our model are vivid (\cref{fig:competitor-style}, middle) and meaningful (\cref{fig:competitor-style}, left, right), the gestures of StyleGestures clearly fall into a mean pose with little variation.

The model trained with the TWH dataset produces a nodding motion in sync with an agreement, ``yeah" during the conversation (see the supplementary video).  Overall, all the baselines produce repetitive motions without much contextual relationship with the speech, with limited gesture variations. On the other hand, BodyFormer can generate gestures that are closely correlated to the semantics of the speech and produces a wild range of different motions.


\begin{figure}[t]
	\begin{tabular}{cc}
	\includegraphics[width=0.475\linewidth]{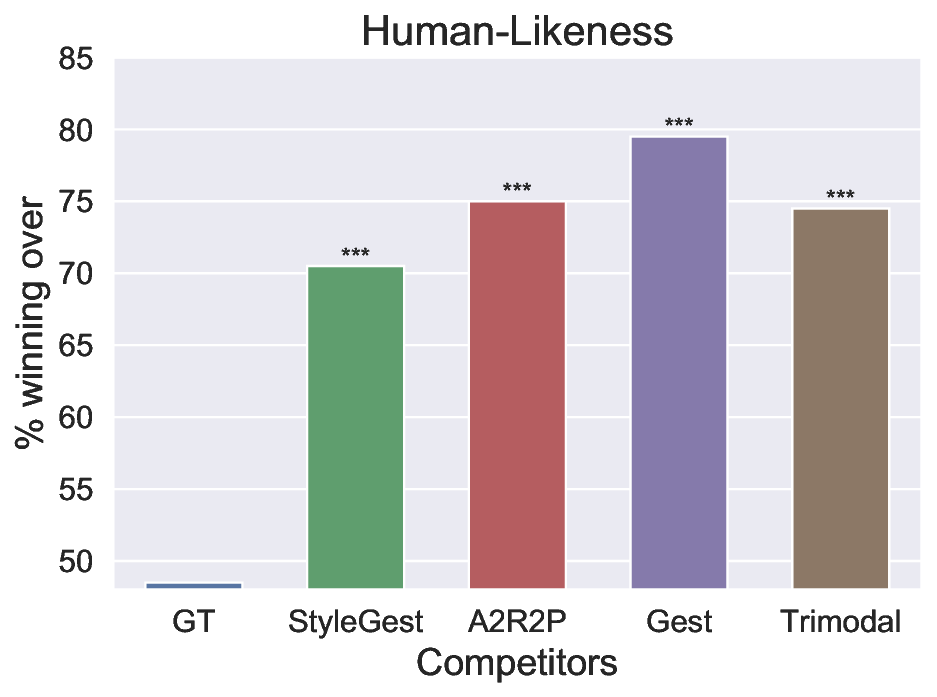}	&
	\includegraphics[width=0.475\linewidth]{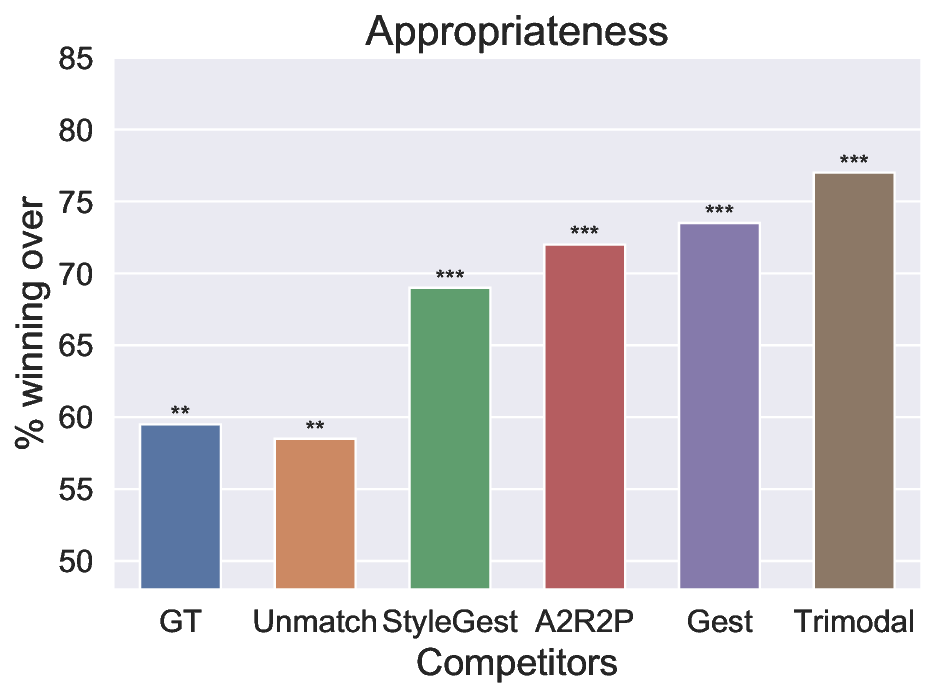}	\\
	(a)&(b)
	\end{tabular}
 	\caption{The pairwise winning proportion of the competitors for (a) Human-likeness and (b) Appropriateness on the Trinity dataset. Asterisks indicate significant effects (**:p<0.05, ***:p<0.001). }
	\label{fig:genea}
\end{figure}

\begin{figure}[t]
	\centering
	\begin{tabular}{cc}
	\includegraphics[width=0.475\linewidth]{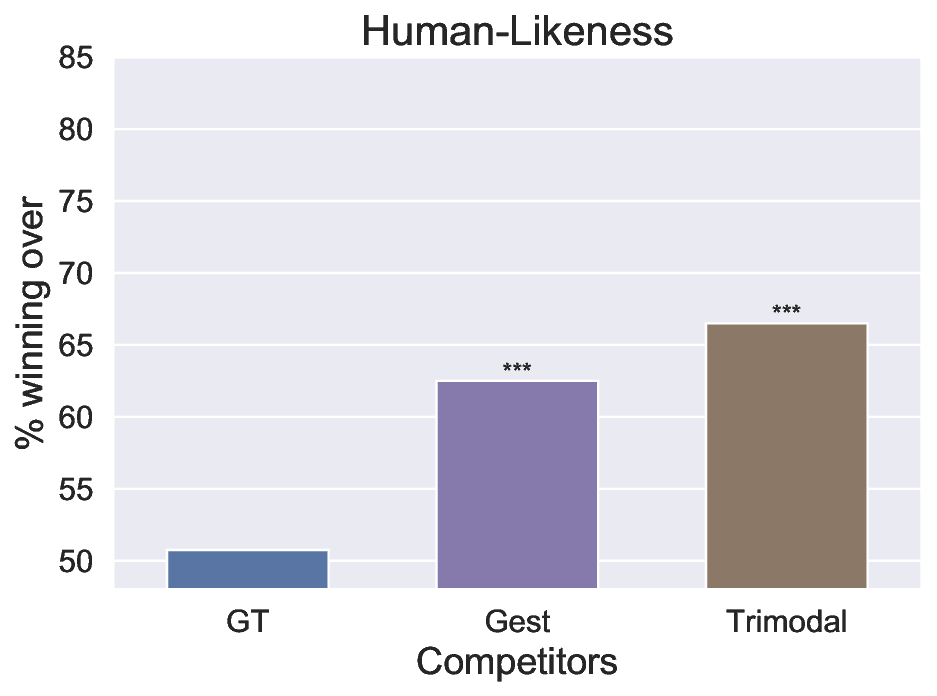}	&
	\includegraphics[width=0.475\linewidth]{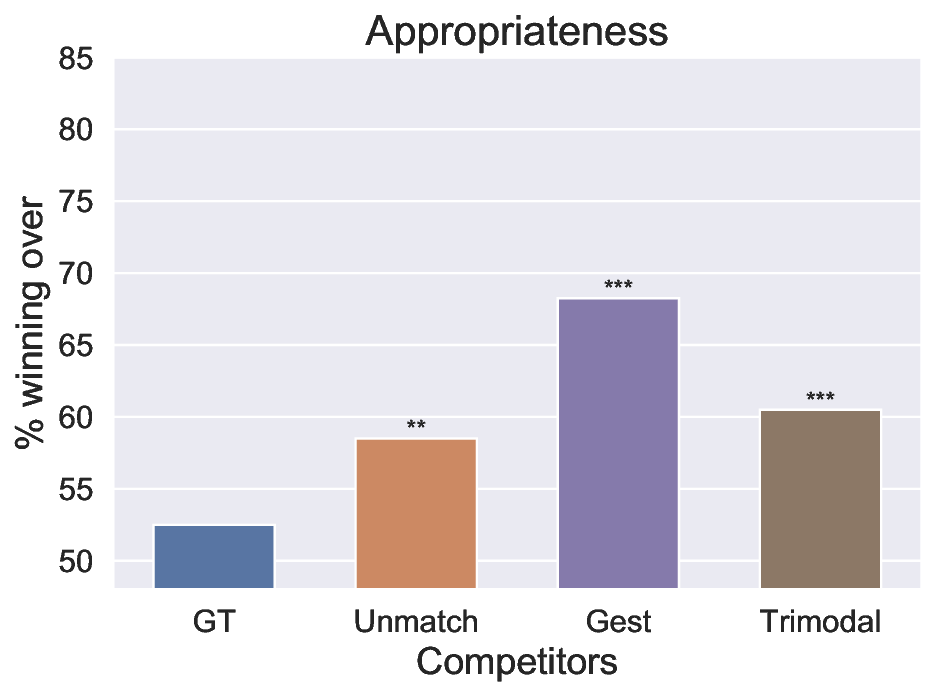}	\\
	(a)&(b)
	\end{tabular}
 	\caption{The pairwise winning proportion of the competitors for (a) Human-likeness and (b) Appropriateness, on the TWH Dataset. Asterisks indicate significant effects (**:p<0.05, ***:p<0.001). }
	\label{fig:twh_user_study}
\end{figure}

\begin{table*}
	\centering
	\caption{The mean and standard deviation about the sum of the norm of velocity. } 
	\label{tab:mode_ablation}
	\begin{tabular}{|c|c|c|c|c|c|c|}
		\hline
		 & \multicolumn{3}{c|}{Trinity}& \multicolumn{3}{c|}{Talking with Hands}\\
		 \hline
		Speaking mode &  NS & SS & LS &  NS & SS & LS \\
		\hline
		Ground Truth  & $38.97 \pm 19.66$ & $40.37 \pm 22.59$ & $46.83 \pm 24.63$ & $38.19 \pm 16.20$ & $44.90 \pm 22.50$ & $53.32 \pm 28.35$ \\
	\hline
	Our Proposed Embedding Layer  & $32.97 \pm 11.09$ & $42.53 \pm 13.60$ & $40.69 \pm 11.86$ & $27.32 \pm 07.52$ & $30.26 \pm 07.62$ & $40.09 \pm 13.14$\\
	\hline
	Global Positional Embedding  & $26.33 \pm 14.20$ & $23.80 \pm 10.99$ & $23.75 \pm 09.77$ & $19.34 \pm 06.68$ & $19.31 \pm 04.58$ & $28.91 \pm 10.93$ \\
	\hline
	w/o Motion Decoder Pre-training  & $23.42 \pm 11.38$ & $23.27 \pm 07.63$ & $28.41 \pm 10.28$ & $26.26 \pm 07.09 $ & $29.70 \pm 06.21$ & $38.13 \pm 12.32$
	\\ 
	\hline
	\end{tabular}
 
\end{table*}

\subsection{User Study}

We compare our method with the state-of-the-art methods through the user study. We adopt the same criteria by \cite{DBLP:journals/cgf/AlexandersonHKB20} for the subjective evaluation: The \emph{Human-likeness} is a measure of ``how human-like the gesture appears'' without considering the correlation between the speech and gesture. The \emph{appropriateness} measures ``how appropriate are the gestures for the speech?''. The participants are expected to evaluate the tightness between motion and speech rather than the motion quality \cite{kucherenko2021large}. 

For the user study, we do an A-B testing where the user labels the winner for every pairwise video, which is a combination of videos of a character controlled by our method and that by the other competitors. We render the videos of the character with Blender. For the methods whose outputs are joint positions or directional vectors, we do the inverse kinematics to recover the joint rotations. 

The subjective evaluation results are collected via the Amazon Mturk platform\footnote{\url{https://www.mturk.com/worker}}, \new{where the workers were hired with at least a historical accept rate of 98\%}. More specifically, for the Trinity experiment, we obtained 40 non-overlapping 15-second audio clips, and for each clip, we created 11 pairwise videos: ten for our model versus the baseline and one attention check for ground truth versus a still pose. Failing the attention check would redo the whole evaluation until pass.  The order and pairs of the videos were randomized, and each clip was evaluated by five workers, resulting in 2200 comparisons and 152 unique participants. \new{The same procedure was followed for the TWH experiment, where we acquired 80 non-repeated audio clips with six pairwise videos each, resulting in 2400 comparisons and 302 unique participants.}

We show the winning proportion in \cref{fig:genea}. From the results on Trinity Speech-Gesture Dataset, we can see that users prefer our method over other competitors (StyleGestures, Aud2Repr2Pos, Gesticulator and Trimodal) in terms of Appropriateness and Human-likeness. Among the four competitors, StyleGesture performs the best. Turkers perceive the gestures generated by our method more human-like and more appropriate than StyleGesture (Human-likenss: 70.5$\%$, Appropriateness: 69.0$\%$). We notice that Aud2Repr2Pos, Gesticulator and Trimodal generate relatively static gestures. 

We infer that those CNN and RNN-based gesture generators are restricted to training on short sequence generation, which may potentially lose long-term information due to the convolutional filter's limited range or the recurrent module's inevitable forgetting. The transformer is suitable for automatically discovering both short and long dependencies, allowing it to produce contextually meaningful long motion sequences (in our case, 15 seconds). As shown in fig.~\ref{fig:trinity_heatmap}, the transformer can capture dynamic and longer dependencies across sequences which helps improve the motion quality. 
While StyleGestures performs slightly better, the range of motion is relatively small, and some of the body motions are not semantically correlated. One possible reason is that StyleGestures does not consider different motion speeds in different speaking modes, whereas our method can capture such variations. Interestingly, when comparing our method to the ground truth, our method ranks slightly better than the ground truth on the appropriateness (Human-likeness:  48.5$\%$, Appropriateness: 59.5$\%$). In addition, according to the \cref{fig:genea} and \cref{fig:twh_user_study}, the paired t-tests further reveal that our model performed significantly better than all baselines (p<0.001) except the ground truth, both for the Trinity and TWH datasets. Ours was also significantly better than ground truth (p<0.05) for appropriateness on the Trinity dataset.



\begin{table}
	\centering
	\caption{Trinity Quantitative Metrics}
	\label{tab:quant_metrics_trinity}
	\begin{tabular}{|c|c|c|c|c|c|c|}
		\hline
		&  Ours & w/o D.P & Tri & StyleGest &  Gest & A2R2P \\
		\hline
		MAJE & $70.05 $ & $71.92$ & $126.71$ & $107.95$ & $86.04$ & $125.50$  \\
		\hline
		FGD  & $9.66 $ & $10.83$ & $254.90$ & $20.81$ & $51.53$ & $346.50$ \\
		\hline
	\end{tabular}
\end{table}

\begin{table}
	\centering
	\caption{Talking with Hands Quantitative Metrics}
	\label{tab:quant_metrics_twh}
	\begin{tabular}{|c|c|c|c|c|}
		\hline
		&  Ours & w/o D.P & Tri & Gest \\
		\hline
		MAJE & $81.66$ & $82.79$ & $187.47$ & $157.28$ \\
		\hline
		FGD  & $12.12 $ & $16.75$ & $47.02$ & $55.26$ \\
		\hline
	\end{tabular}
\end{table}


Apart from the qualitative evaluation, we adopt two commonly used quantitative evaluation metrics of motion synthesis tasks to compare all methods: Mean Absolute Joint Error (MAJE) and Frechet Gesture Distance (FGD)~\cite{rhythm_sigasia2022_ao, 10.1145/3414685.3417838}. The MAJE helps to measure the closeness of joint positions between the natural human and the generated motion, and the FGD quantifies the divergence of distributions under the Gaussian assumption. Tables~\ref{tab:quant_metrics_trinity} and \ref{tab:quant_metrics_twh} summarize performance of all the methods on the Trinity and TWH datasets, respectively. Not only does the proposed approach consistently outperform the competitors on the MAJE, but it also shows a more plausible generated motion in terms of FGD.

\subsection{Ablation Study}
\begin{figure}
\includegraphics[width=\linewidth]{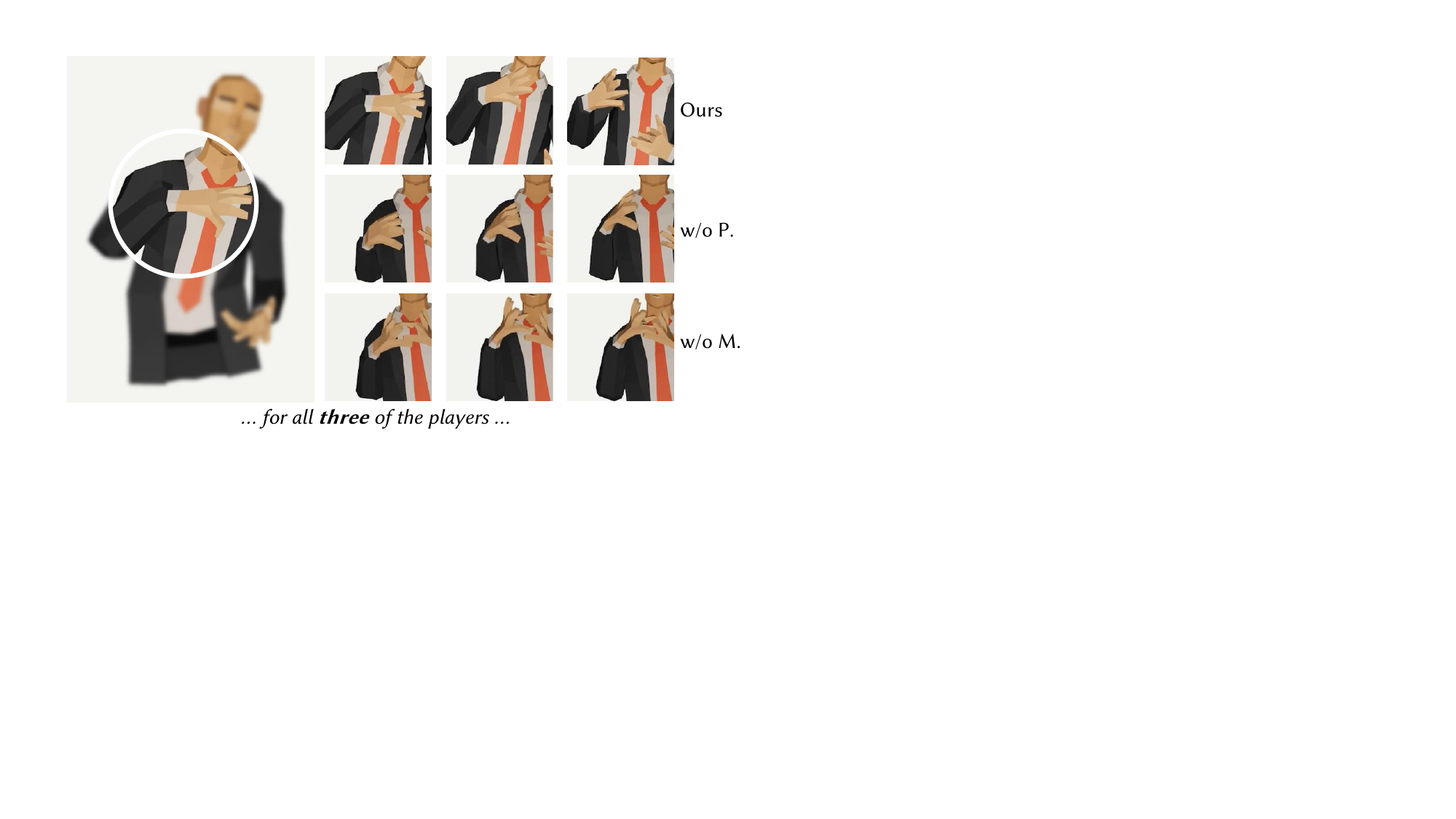}
\caption{\label{fig:ablation_three}The character delivers the word \textit{\textbf{three}} perfectly in our model, while the gesture does not match the onset of \textit{\textbf{three}} without pre-training and his hands remain static without MPE}
\end{figure}

\begin{figure*}
\includegraphics[width=\textwidth]{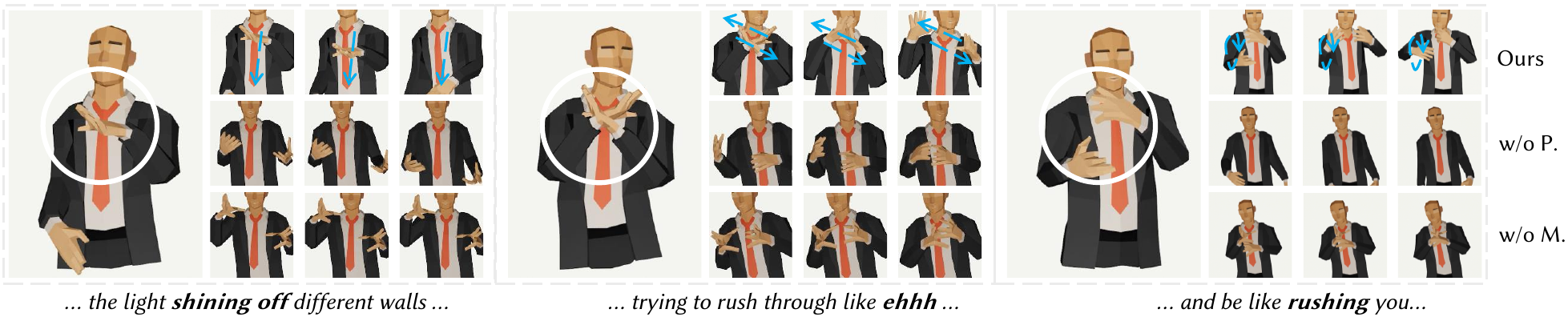}
\caption{\label{fig:ablation}\textbf{Left:} When the speaker says \textit{\textbf{shining off}}, the character raises his right hand up and then put it down in front of his chest in our model, trying to emphasize the verb, while his right-hand swings meaninglessly without pre-training and the motion is nearly static without MPE. \textbf{Middle:} In our model, his hands stop suddenly as the speaker utters the hesitation marker \textit{\textbf{ehh}}. But the pause does not show without pre-training and is lagged behind the speech without MPE. \textbf{Right}: The hands move quickly to express the keyword \textit\textbf{{rushing}} in our model. The hands stay in the rest pose without pre-training and move slowly without MPE. (See the supplementary video for dynamic comparisons.)}
\end{figure*}

\begin{figure}[t]
	\centering
	\includegraphics[width=\linewidth]{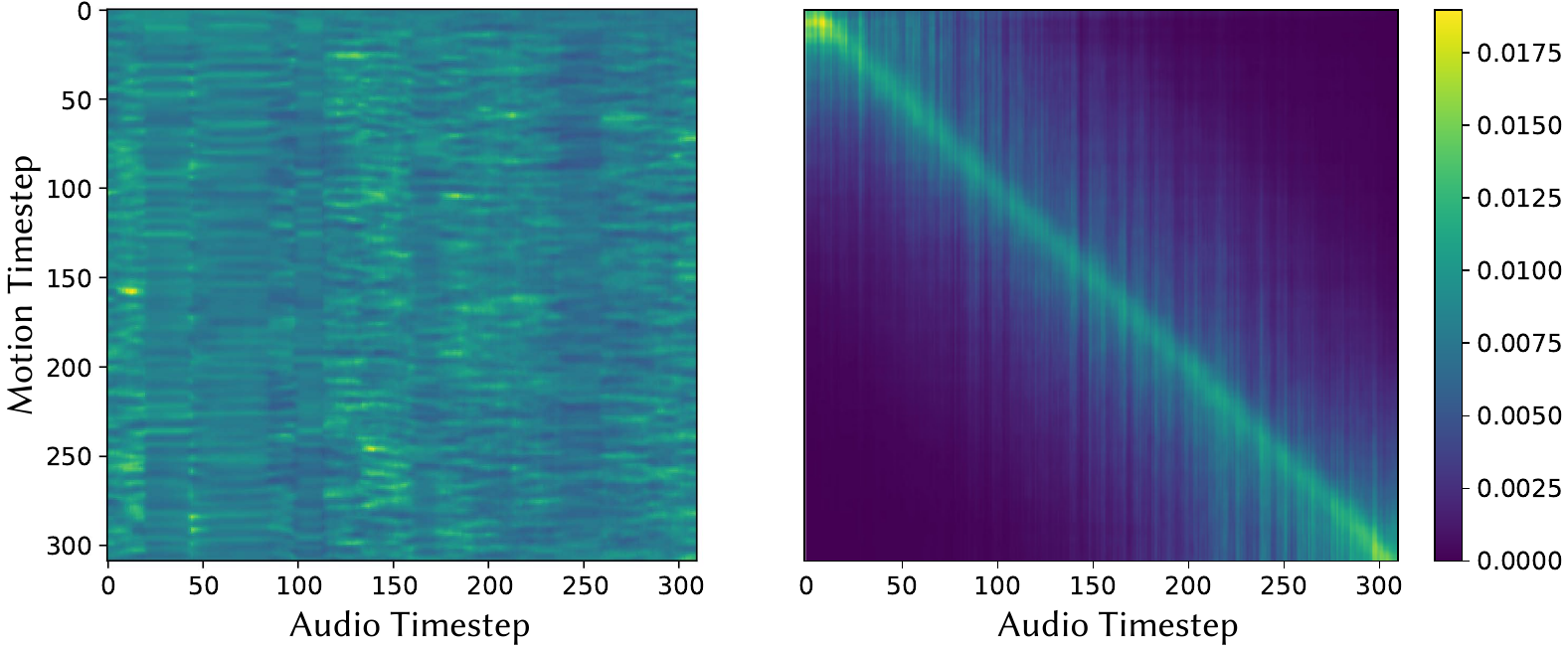}
 	\caption{Attention heatmap average across all test sequences without repetitions where (a) w/o P (b) Ours.}
	\label{fig:trinity_heatmap}
\end{figure}

\begin{figure}[t]
	\centering
	\begin{tabular}{cc}
        \includegraphics[width=0.475\linewidth]{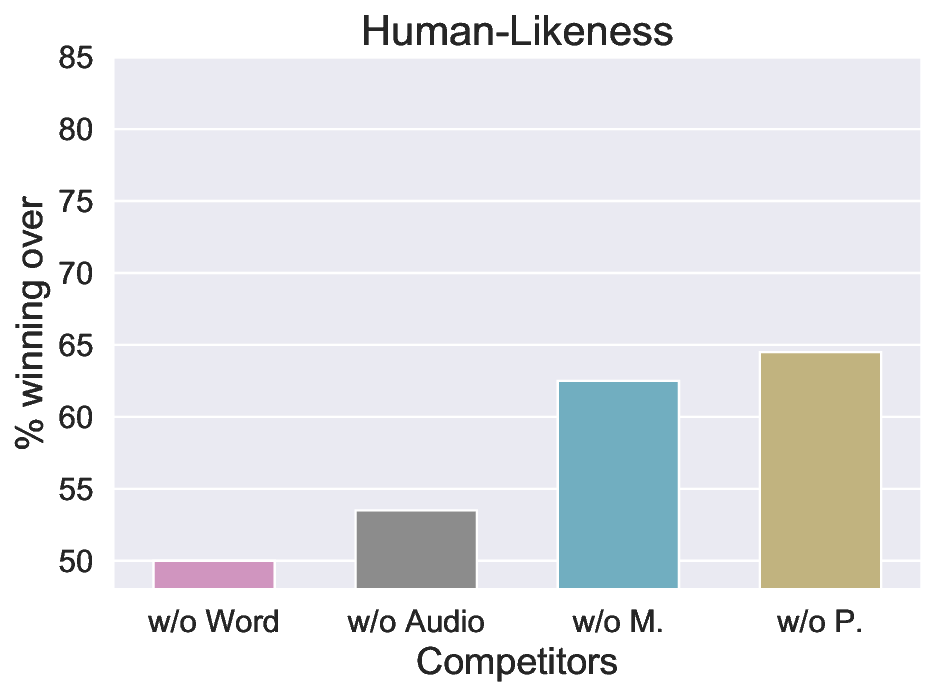}	&
	\includegraphics[width=0.475\linewidth]{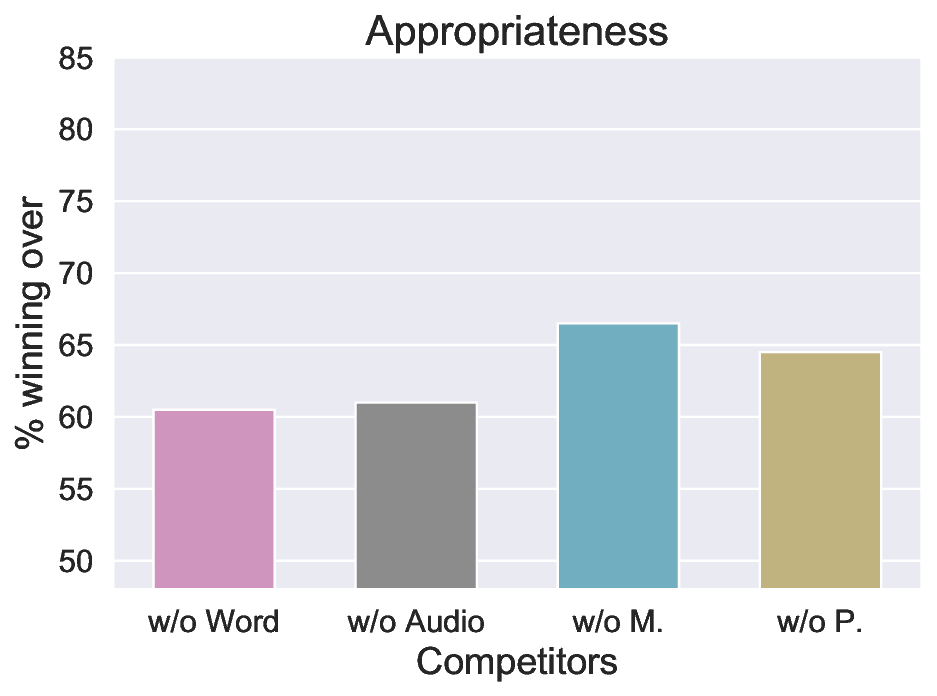}	\\
	(a)&(b)
	\end{tabular}
 	\caption{The pairwise winning proportion of the competitors (a) Human-likeness and (b) Appropriateness, on Trinity Speech-Gesture Dataset.}
	\label{fig:trinit-ab}
\end{figure}

\paragraph*{Intra-modal Pre-training} Since the motion datasets are of relatively small size to train a transformer, appropriate pre-training becomes crucial. To validate our proposed \emph{Intra-modal Pre-training},  we test a version of our model without using this strategy, denoted as ``w/o P''. Figs. \ref{fig:ablation_three} and \ref{fig:ablation} visualize the generated gestures of our model (1st row) and ``w/o P'' (2nd row). Note that in \cref{fig:ablation_three}, while the gesture of ``w/o P'' fails to match the onset of the word ``three'', our model can flawlessly deliver the sign language for ``three.'' This suggests that our model benefits from the intra-modal pre-training. This is also supported by the attention map (\cref{fig:trinity_heatmap}) showing that our model can automatically learn the correspondence between the contextualized speech representations and the motions. The user study results (\cref{fig:trinit-ab}) also suggest the effectiveness of the intra-modal pre-training (Ours vs. ``w/o P'' - Human-likeness: 64.5$\%$, Appropriateness: 64.5$\%$). The readers are referred to the supplementary video for the animation comparison.

To further analyze the effectiveness of the intra-modal pre-training, we provide an ablation study to validate the necessity of motion decoder pre-training. Though tables~\ref{tab:quant_metrics_trinity} and \ref{tab:quant_metrics_twh} show that with and without motion decoder pre-training ``w/o D.P'' has similar MAJE and FGD, table \ref{tab:mode_ablation} shows a significant difference in cumulative velocity. With the motion decoder pre-training, BodyFormer has a similar distribution to the ground truth on the three speaking modes in the Trinity dataset. The norm of velocity decreases significantly without the motion decoder pre-training, which indicates a loss of gesture quality.  We argue that the motion decoder pre-training can learn non-verbal motion to help generate active and fast-changing gestures.

\paragraph*{Mode Position Embedding} We present the ablation study on the mode position embedding. The result of our model with the mode position embedding layer removed is shown in \cref{fig:ablation}, ``w/o M'' (row 3). The user study results (\cref{fig:trinit-ab}) show that removing the position embedding leads to less human-like and appropriate body gestures (Ours vs. ``w/o M'' - Human-likeness: 62.5$\%$, Appropriateness: 66.5$\%$).

In addition, we also compare the motion variations between the proposed embedding and the global positional embedding. We measure the cumulative sum of L2 norm of velocity 
for every mode. 
As shown in the \cref{tab:mode_ablation}, our proposed embedding has a higher cumulative sum of the L2 norm of velocity than the global positional embedding and is comparable with the ground truth on the Trinity dataset. Though the statistics of our proposed embedding still have a gap between the ground truth on the TWH dataset, our motion speed is still faster than the global positional embedding.
This shows that the global positional embedding potentially over-smooths generated motion by averaging the motions across all modes.

\paragraph*{Usefulness of Audio and Words}
We provide an ablation study of the usefulness of audio and words. According to \cref{fig:trinit-ab}, our method scores better than ``w/o Word'' and ``w/o Audio'' on both Human-likeness and Appropriateness aspects. 
Intuitively, the mel-spectrogram feature is expected to capture the synchronization between audio and motion, resulting in more natural-looking generated motion. However, the BERT feature alone may not capture this information. As a result, combining both the audio and word features in our method results in superior performance compared to using only the audio feature (``w/o Word'') or only the word feature (``w/o Audio''). This is because using only audio without semantic information may cause the model to struggle with generating complex motion. High-level context information provided by the BERT feature can complement the audio, resulting in better overall performance.
We provide a detailed comparison in our video. Readers are referred to the video for further comparison.

\section{Limitations}
Though we provide a way to measure the performance of speech-to-body motion synthesis by comparing the average motion speed for each mode, it is still an open question of how to evaluate the quality of the synthesized motion. Our system is a seq2seq framework that uses future text and speech information for generating the gesture. This could be an issue for real-time applications such as remote meetings.  A possible solution can be to predict the rest of the sentences from the speech up to now and input that into the system for motion synthesis.

\section{Conclusion and Future work}
In this paper, we present \textit{BodyFormer}, a method to synthesize body gestures from speech data using a variational transformer model. The system produces realistic and vivid motion and outperforms existing models both in quantitative and qualitative evaluation. We are interested in further improving the model through a better evaluation metric and a training scheme that can learn from a wider range of data. We are also interested in constructing a multi-modality model that also learns the facial expressions and gaze for metaverse applications.

\begin{acks}
This research is supported by 
Meta Reality Labs,   
Innovation and Technology Commission
(Ref:ITS/319/21FP) and Research Grant Council (Ref:  
17210222), Hong Kong.   
\end{acks}

\bibliographystyle{ACM-Reference-Format}
\bibliography{egbib}
\appendix 
\section{Automatic Mode Labeling}
\label{apdx_mode}
\new{
The mode labeling is determined by the time length of each speech. 
\begin{itemize}
    \item \textbf{Long Speaking}: the speech longer than 2 seconds.
    \item \textbf{Short Speaking}: the speech shorter than 2 seconds.
    \item \textbf{Non Speaking}: no word recognized by the Google ASR tools.
\end{itemize}
}
\section{Baseline Implementations}
\begin{figure}[H]
\centering
\includegraphics[width=0.8\linewidth]{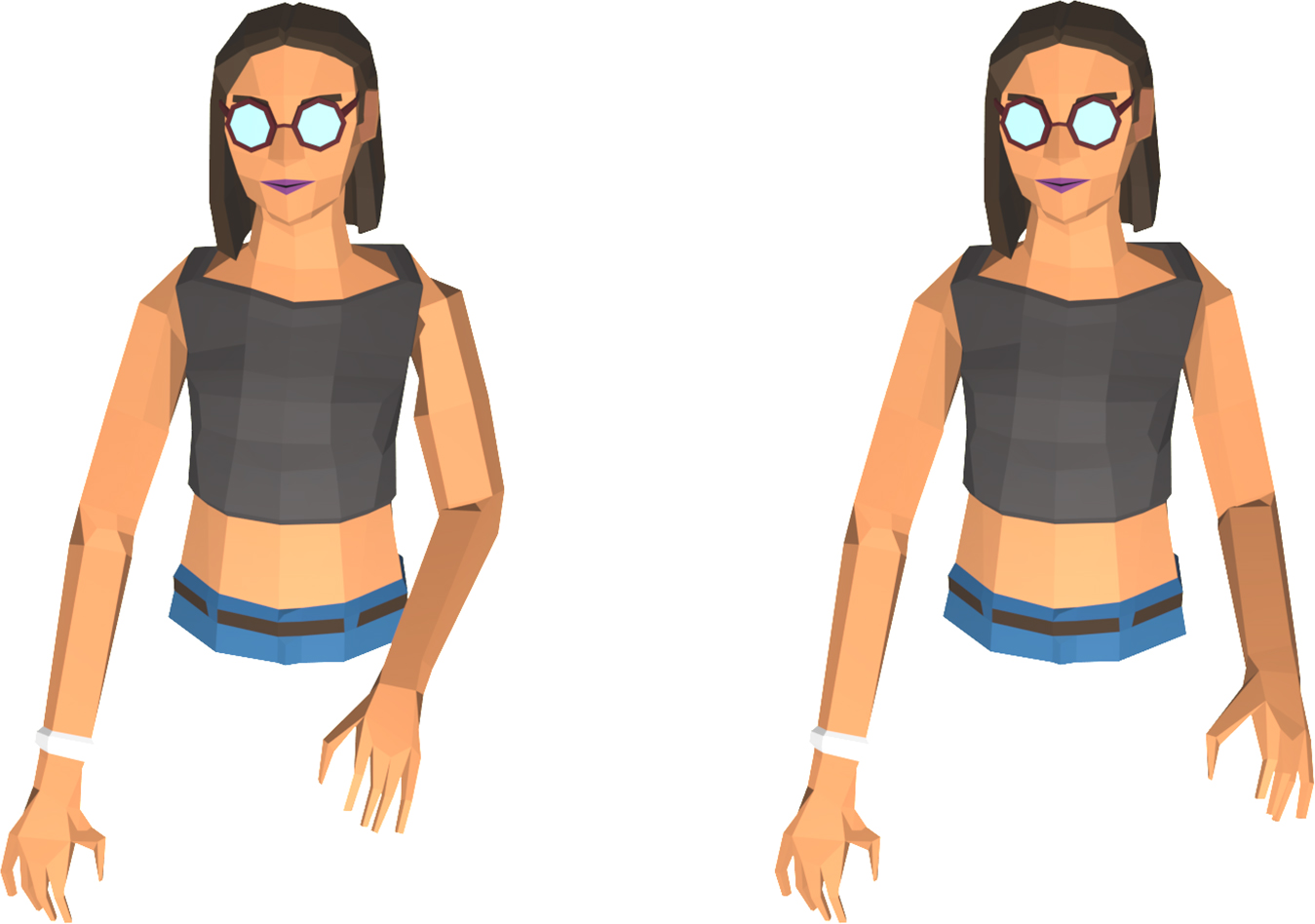}
\caption{\new{\textbf{Left:} A motion frame generated by Gesticulator on TWH dataset. Note that the left shoulder is wrongly rotated due to the error of inverse rigging. \textbf{Right:} The same frame after the manual fix. As this error is consistent between different frames, we fix it by simply applying a rotational offset on all frames.  }}
\label{fig:manual_fix}
\end{figure}

For baselines that output the global position of each joint, we apply Inverse-Kinematics (IK) to recover the joint rotations.
For the TWH dataset, we compare BodyFormer with Trimodal and Gesticulator.
\new{
Since TWH has a non-standard T-pose, direct IK on the TWH skeleton leads to failure. To map the baseline motions to the TWH skeleton, we create a proxy skeleton with a standard T-pose and apply IK to it. Then, we utilize  the Rokoko blender plugin~\footnote{\new{https://www.rokoko.com/integrations/blender}} to transfer motion to the TWH skeleton. We also tackle motion defects from IK by applying rotation offsets to problematic joints. See \cref{fig:manual_fix} for an illustration.}

\new{We use the provided codes of each author to implement the baselines. As none of them considered finger motions, we adopted a default pose to prevent finger collapse. All methods are rendered by Blender under the same setting of 20 fps, except for Trimodal, which is rendered in 15 fps to follow their original implementation.}

\section{Example Output of User Study}

\begin{figure}[H]
    \centering
    \includegraphics[width=0.9\linewidth]{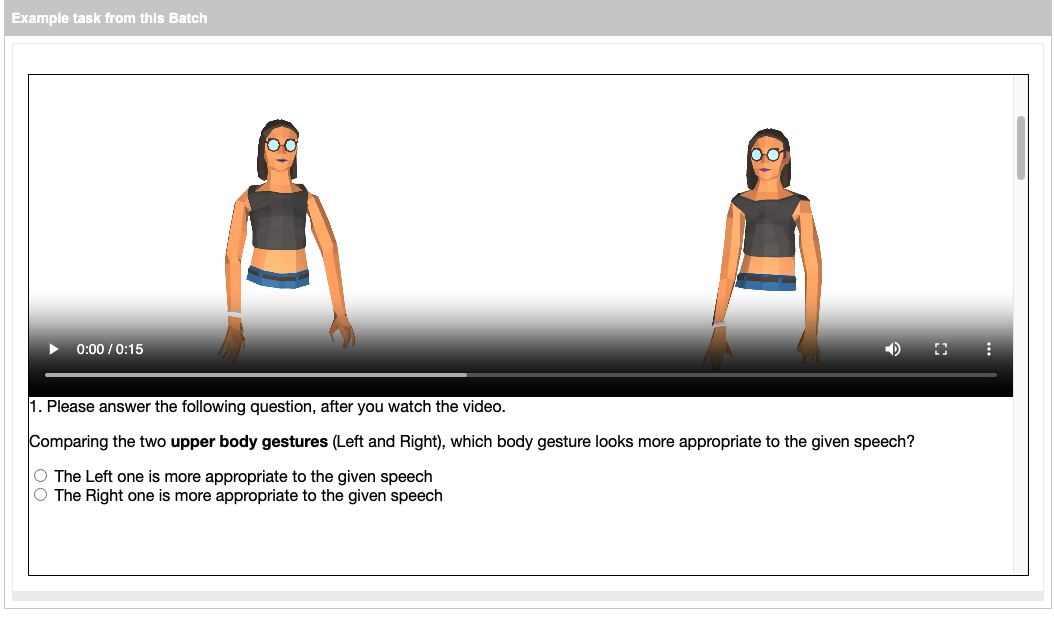}
    \caption{The example of the user study}
    \label{fig:appendix_userstudy}
\end{figure}

\new{As shown in the \cref{fig:appendix_userstudy}, an example output of our user study focused on appropriateness. A question that the participants were asked was: "Which body gesture looks more appropriate to the given speech?" To answer this question, participants could click the "play" button to listen to the audio clip and then select the appropriate radio button before submitting their user study. We also conducted a similar study focused on human-likeness, which used the same framework but without audio.}

\end{document}